# Integrating Supervised Extractive and Generative Language Models for Suicide Risk Evidence Summarization

**Rika Tanaka**[*]       **Yusuke Fukazawa**[*]
Graduate Degree Program of Applied Data Sciences
Sophia University Graduate School
Tokyo, JAPAN

## Abstract

We propose a method that integrates supervised extractive and generative language models for providing supporting evidence of suicide risk in the CLPsych 2024 shared task. Our approach comprises three steps. Initially, we construct a BERT-based model for estimating sentence-level suicide risk and negative sentiment. Next, we precisely identify high suicide risk sentences by emphasizing elevated probabilities of both suicide risk and negative sentiment. Finally, we integrate generative summaries using the MentaLLaMa framework and extractive summaries from identified high suicide risk sentences and a specialized dictionary of suicidal risk words. SophiaADS, our team, achieved 1st place for highlight extraction and ranked 10th for summary generation, both based on recall and consistency metrics, respectively.

## 1 Introduction

Identifying suicide risk from online discussions is crucial problem. The 2018 and 2019 Shared Task at CLPsych posed the task of predicting the level of suicide risk annotated by experts from Reddit posts (Shing et al., 2018; Zirikly et al., 2019).

In the 2024 Shared Task (Chim et al., 2024), the task is further developed to provide supporting evidence about an individual's suicide risk level on the basis of their linguistic content. There are two related subtasks. First subtask is to provide highlights; relevant evidence spans supporting the expert-assigned risk level. Second subtask is to provide evidence summaries which synthesizes the identified evidence into insights that are consistent with human-written summaries.

Two main approaches for text summarization exist: extractive and generative. The extractive approach focuses on selecting significant portions of the original text, often using techniques like sentence extraction and machine learning-based sentence ranking (Ferreira et al., 2013; Aliaksei et al., 2015). In contrast, the generative approach involves creating coherent summaries by understanding the context and meaning of the input text, employing advanced neural network architectures such as Transformer models pre-trained for language understanding and generation (Vaswani et al., 2017; Brown et al., 2019).

The extractive approach excels in selecting crucial sentences based on supervised learning and explicit extraction criteria. In contrast, the generative approach is advantageous for understanding context and generating summaries without the need for explicit guidance. While the extractive approach struggles with the holistic contextual consideration, the generative approach faces challenges in reliably extracting desired evidence through prompt engineering. Consequently, when clear criteria are present, the extractive approach is preferable; however, for generating contextually comprehensive summaries, the generative approach is more suitable.

Given the dual requirements of this year's shared task – identifying high suicide risk sentences and comprehensively considering various aspects of the entire post, including emotions, cognitions, behavior, and motivation – we propose an integrated method combining both extractive and generative approaches. Our contributions include: (1) Developing a BERT (Devlin et al., 2019) based model for sentence-level suicide risk and negative sentiment estimation. (2) Identifying high suicide risk sentences precisely by focusing on elevated probabilities of both suicide risk and negative sentiment. (3) Integrating generative summaries using the MentaLLaMa (Yang et al., 2023) framework and extractive summaries from identified high-risk sentences and a specialized suicidal risk words dictionary. The following sections detail our proposed method, results, and conclusion.

---

[*] Both authors contributed equally, with Ms. Tanaka on MentaLLaMa and Prof. Fukazawa on BERT.

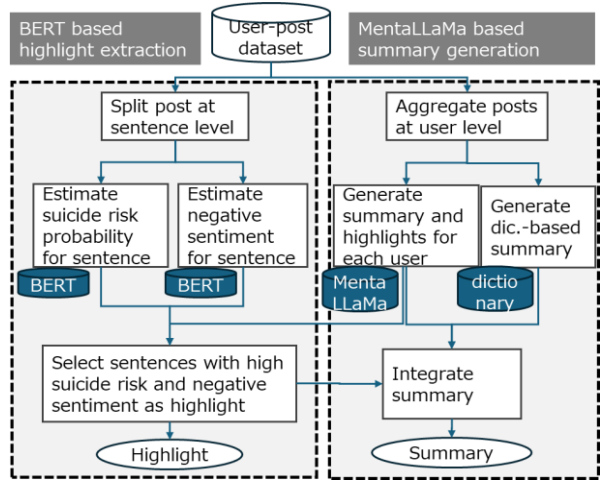

Figure 1: Overall process of proposed model.

## 2  Proposed method

The proposed method, outlined in Fig. 1, comprises two parts: highlight extraction and summary generation. To identify sentences indicating suicide risk, we employ a supervised extractive approach, leveraging BERT's fine-tuning capabilities for enhanced contextual understanding. Our model, fine-tuned on BERT, estimates suicide risk and negative sentiment at the sentence level. For summary generation, we combine extractive and generative approaches. Extractive summaries are crafted using patterns derived from high suicide risk sentences and associated keywords. Generative summaries are produced using MentaLLaMa. The overall summary is an integration of both approaches.

In the following, we detailed sentence level suicide risk classification, sentiment classification, highlight extraction, and summary generation.

### 2.1  Sentence level suicide risk classification

#### 2.1.1  Extraction level

The decision of extraction level, be it word, phrase, sentence, or paragraph, is crucial. To capture effective contextual information, a minimum consideration of the phrase level is necessary. Examining words around the phrase is vital for strengthening the evidentiary basis for suicide risk. However, paragraph-level extraction introduces the risk of irrelevant context, prompting our choice of sentence-level extraction in this study. Each post is divided into sentences by punctuation marks (.,!?;:).

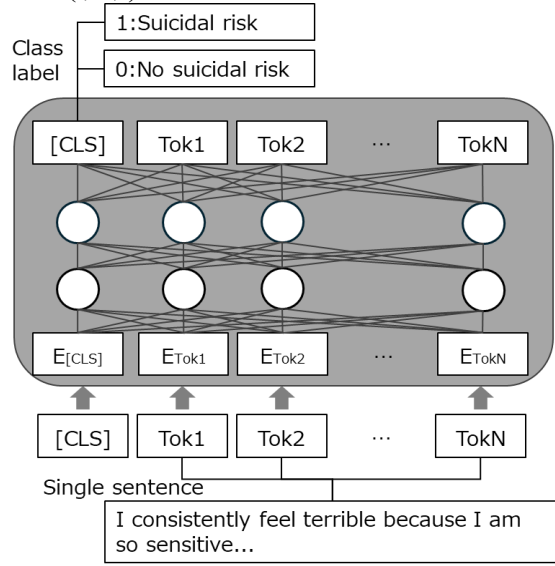

Figure 2: BERT finetuning for sentence level assessment of suicide risk.

#### 2.1.2  BERT finetuning

To assess sentence level suicide risk estimation, we adopt BERT finetuning approach. We prepared training data for finetuning by collecting sentences that refer to suicide in direct expressions. We found that suicide risk sentences contained characteristic phrases as shown in Appendix A. We collected sentences containing the phrases in Appendix A as suicide risk sentences. As a result, the number of sentences containing those phrases was 557 (label 1), and the number of other sentences (label 0) was 31,428. In order to balance the number of labels, we down sampled the one with label 0. As a result, we acquired the training data (label 1: 449, label 0: 412) and validation data (label 1: 108, label 0: 115).

We utilize the BERT model[1] as depicted in Fig. 2. The finetuning process involves inserting a [CLS] token at the text's start, tokenizing the data, and using the Transformer architecture to abstract sentence representations (E) for each token. The E[CLS] representation of the [CLS] token captures the sentence's meaning and context. A fully-connected layer (classifier) applies a softmax function to E[CLS] to generate class probabilities. Both the embedded representation and the classifier's parameters are adjusted to predict the golden labels (suicide risk: 1, no suicide risk: 0) for the input text.

---

[1]  https://huggingface.co/bert-base-uncased

The parameters used for training are as follows; The batch size utilized during the training phase is set to 8. The learning rate, a crucial hyperparameter governing the model's weight updates during training is 2e-5. Warmup ratio controlling the initial gradual increase of the learning rate is 0.1. The evaluation metric utilized to determine the best model is accuracy.

Fig. 3 displays the learning curve, with training halted at epoch=50 for presumed convergence. Using the model with the highest accuracy (0.996), we predicted labels for all sentences, obtaining suicide risk labels and associated probabilities. The results revealed the model's capability to detect previously undetected phrases, such as "Dying is the only way to make it better" and "fall asleep and never wake up," which were not identified by the phrases listed in Appendix A.

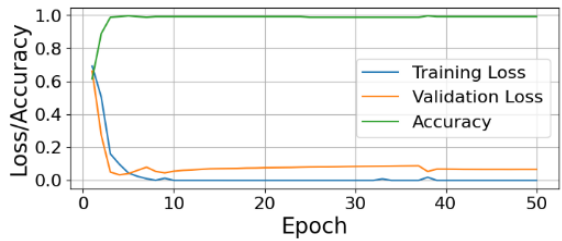

Figure 3: Learning curve of finetuning

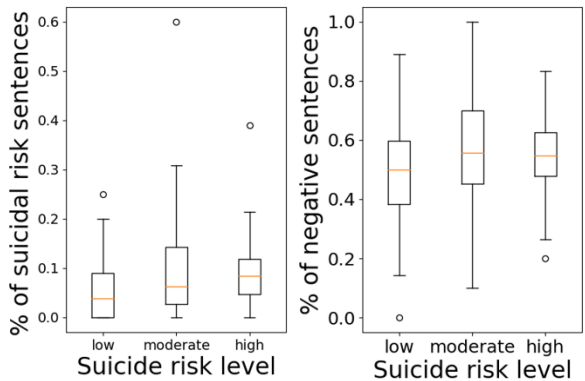

Figure 4: Boxplots of the number of suicide risk / negative sentiment sentences across user's post for each suicide risk level.

### 2.1.3 Correlation to suicide risk level

The data provided are flagged by experts on the levels of suicide risk: low, moderate, and high risk. In Fig.4 (left), we examine the correlation between ratio of suicide risk sentences in user's post and suicide risk level of corresponding user. We can see that the ratio of suicide risk sentences increases as the levels judged by experts increases. Consequently, the sentences classified as high suicidal risk demonstrates a high potential for the evidence of suicide risk.

### 2.2 Sentiment classification

In this section, we extract sentences with high suicide risk in terms of sentiment. The link between negative emotions and suicide risk is debated. Monselise et al. observed a slight increase in the proportion of negative sentiments before and just after the first suicidal ideation in Reddit user posts (Monselise et al., 2022). In contrast, Gaur et al. found no significant variation in sentiment and emotions across suicide risk severity levels using AFINN and LabMT in C-SSRS (Gaur et al., 2021).

We classify sentence into negative, neutral or positive sentiments using sentiment classification. We used the finetuned BERT model[2], which is currently the latest model trained on short sentences of X posts (Loureiro et al., 2022). X is a social network platform that allows users to post short sentences about their daily events and thoughts. Reddit, on the other hand, is a social network platform where users can post long sentences about their problems and troubles. Although the contents of X and Reddit are different, in this study, we decompose the long sentences of Reddit and perform sentiment classification on a sentence level. For this reason, we used fine-tuned BERT with the X data as the teacher data for the classification.

In Fig.4 (right), we examine the correlation between ratio of negative sentiment sentences in user's post and suicide risk level of corresponding user. We can see that the ratio of negative sentiment sentences increases as the levels judged by experts increases from low to moderate. Consequently, sentences with negative sentiment may be evidence of suicide risk.

### 2.3 Highlight extraction

First, we select all sentences classified as high suicide risk as highlights. Then, we sort sentences in order of probability of negative sentiment and get sentences as highlights from the top to the bottom until the total word count is within 300. If still short of 300 words, we add highlights by MentaLLaMa[3], which is a LLaMA2 (Touvron et al., 2023) finetuned by large collection of social media data related to mental health. We make a query prompt to MentaLLaMa like "The text below implies a risk of suicide. Extract only the necessary and sufficient phrases and keywords indicating the risk exactly as they appear in the original text. Present the extracted words in a list format, separated by commas." with the post aggregated on a per-user basis.

---

[2] https://huggingface.co/cardiffnlp/twitter-roberta-base-sentiment-latest

We observe that the format of output was unstable as there were a mixture of asterisks, numbering, and comma-separated lists. Therefore, instead of parsing the output, we created all the possible phrase candidates consisting of continuous three or more words from the output text. Then we select the sentence of user post that included one of phrase candidates as highlights.

We used the tokenizer to encode the input text without adding special tokens. For text generation, we set the `max_length` parameter to 1024 tokens, limiting the output size. Additionally, `max_new_token` was set to 128 tokens, controlling the number of newly generated tokens. To enhance text diversity, we activated `do_sample`, enabling random sampling. Temperature and repetition penalty were not adjusted.

## 2.4 Summary generation

Our summary consists of 4 parts as shown in Fig. 5. First, we create the opening summary about the level judged by experts. For low suicide risk user, we output "This person is at low risk of suicide."; for moderate suicide risk user, "This person is at moderate risk of suicide."; and for high suicide risk user, "This person is at high risk of suicide.".

Second, we generate a rule-based summary using the number of sentences classified as high suicide risk across multiple posts by a user. When the number of sentences is 1, we output "This person made a post implying suicide.", when the number of sentences is 2, we output "This person made multiple posts implying suicide.", and when the number of sentences is more than 3, output "This person made lots of posts implying suicide.".

Third, we also generate a dictionary-based summary by collecting important phrases leading to suicide ideation across multiple posts by a user.

The phrases are shown in Appendix B. We define those phrases from several websites on suicide feelings. We generate the summary by connecting prefixes and phrases. We also do same procedure for phrases defined in Appendix A. In this case, we use "This person implies suicide such as" as prefix.

Fourth, we generate summaries using MentaLLaMA. We employ a query prompt "Please summarize the next post in 300 words" with user-aggregated posts. The well-crafted output summaries, capturing user behavior and context, are used as-is.

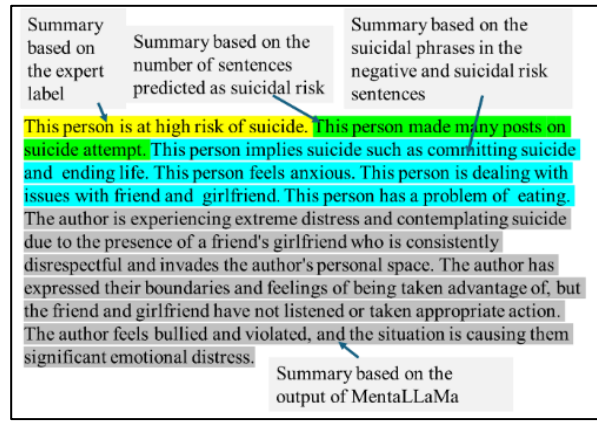

Figure 5: The contents of summary.

## 3 Results

Organizers assess submitted highlights based on recall and precision, with recall measuring gold highlight prediction using BERT-score semantic similarity (Zhang et al., 2019). Precision gauges the quality of predicted supporting evidence. Summarized evidence is evaluated for consistency, indicating the absence of contradiction by calculating the probability of it conflicting with the gold summary. Further details can be found in the paper (Chim et al., 2024).

Two highlight submission patterns were employed - one using only suicide risk classification and the other combining suicide risk classification, sentiment classification, and MentaLLaMa. Table 1 shows results for both patterns. The former achieved the highest precision among all teams, and the latter attained the highest recall among all teams. This underscores the effectiveness of sentence-level suicide extraction for evidence extraction. Sentence-level sentiment classification and MentaLLaMa-based highlight extraction complement in covering additional

---

evidence of suicide risk. Further analysis is provided in the next section.

In the summary generation subtask, it achieved the 10th rank with a consistency metric of 0.944. The lower score is attributed to two reasons: insufficient attention to consistency when integrating multiple summaries and the absence of prompt engineering to incorporate shared task background, relevant aspects, and evaluation metrics into the prompts, despite using simple prompts.

|  | Recall | Precision |
|---|---|---|
| Suicide risk classification | 0.912 | **0.919** |
| +Sentiment classification +MentaLLaMa | **0.944** | 0.906 |

Table 1: Results of highlight extraction subtask for two submission patterns.

## 3.1 Analysis on highlight extraction

For every submitted highlight, we received the semantic similarity between the golden highlight as precision calculated by BERT-Score. We analyzed the correlation between precision and predicted suicide risk/negative sentiment probability for each highlight. Figs. 6 and 7 show the average suicide risk and negative sentiment probabilities for highlight precision. They also display the percentage of highlights with a suicide risk probability of 0.9 or higher and negative sentiment probability of 0.9 or higher. Fig. 6 indicates a correlation between suicide risk probability and precision as evaluated by the golden highlight. In contrast, Fig. 7 shows no correlation between negative sentiment probability and precision as assessed by the golden highlight. This suggests that while sentence-level suicide risk assessment significantly contributes to precise suicide risk evidence highlight extraction, negative sentiment classification does not.

Table 2 presents highlights with high and low precision. High precision highlights frequently articulate users' suicidal thoughts, consistent with previous studies (Rude et al., 2004; Jamil et al., 2017). On the other hand, low precision highlights discuss suicide but often lack actual suicide risk. Instances involve discussions about another person's suicide or expressing negativity towards suicide, such as "I'm not about to commit suicide" and "my best friend also tried to kill himself". This misclassification arises from our suicide risk classification model, which utilizes keyword matching. The training data may include denials of suicide or stories about others' suicides unrelated to personal suicide risk.

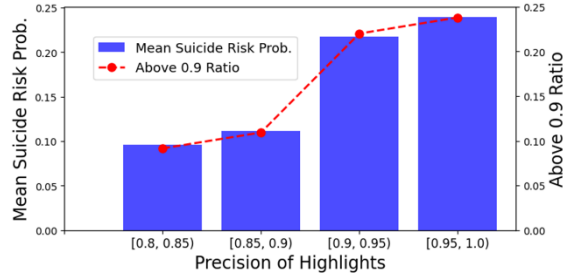

Figure 6: Mean suicide risk probability and above 0.9 ratio vs precision of highlights. We deleted error bar as most of values are close to 0 or 1.

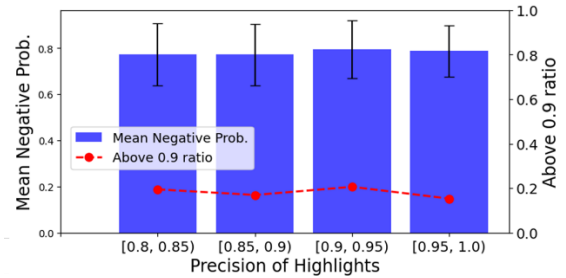

Figure 7: Mean negative sentiment probability and above 0.9 ratio vs precision of highlights.

|  | Phrases |
|---|---|
| Highlights with high precision | I want to die / i am suicidal / I've tried to hang myself two times / I don't know how to stop thinking of suicide |
| Highlights with low precision | I'm not about to commit suicide / I wasnt able to kill myself / My last objection to suicide is that/ losing someone to suicide / I haven't considered actually killing myself / my best friend also tried to kill himself / If you're close to killing yourself |

Table 2: Example highlights that received high and low precision scores. The higher the score, the higher the semantic similarity to gold highlights.

## 4 Conclusion

We proposed integrating supervised extractive LLM (BERT fine-tuned for sentence-level suicide risk extraction) and generative LLM (MentaLLaMa) for summarizing suicide risk evidence. Sentence-level suicide risk assessment achieved the highest precision and recall. Future work will explore replicating these promising results with generative LLMs.

## Limitations

This paper lacks meticulous teacher data creation for suicide risk estimation. Suicide risk has been labeled using keyword matching with the dictionary in Appendix A, potentially introducing noise data like sentences without suicidal thoughts or sentences about others' suicides. To enhance the accuracy of suicide risk classification, manual examination of the training data is necessary. Some participants such as (Sandu et al., 2024) take supervised approach, and we will reference their approaches.

This paper lacks a clear evaluation of why sentence-level surpasses other levels (e.g., word or paragraph) for highlight extraction. In the case of long sentences, there is a possibility that unnecessary parts are highlighted.

The methodology heavily depends on manual design, lacking automation by generative LLMs. While achieving excellent results in highlight extraction, the manual processes hinder scalability and efficiency. Exploring directions to replicate these promising results using generative LLMs is essential, emphasizing the need for automation. Many participating research teams in this shared task such as (Singh et al., 2024) utilized generative LLMs with prompt engineering, and we will reference their approaches.

## Ethical Statement

We have signed a data usage agreement with organizer of CLPsych 2024 ensuring strict adherence to privacy protection and confidentiality. Secure access to the shared task dataset was provided with IRB approval under University of Maryland, College Park protocol 1642625 and approval by the Biomedical and Scientific Research Ethics Committee (BSREC) at the University of Warwick (ethical application reference BSREC 40/19-20). To reinforce the confidentiality of the data, it has been securely stored in an environment accessible exclusively by team members. We excluded API-related LLMs from consideration and focused only on downloadable LLMs.

Prof. Fukazawa, one of the authors, briefed the team on potential mental health impacts during tasks, addressing risks linked to individuals with a history of suicidal thoughts. The team is thoughtfully assembled with members free from mental health concerns, ensuring a supportive and safe work environment.

## Acknowledgments

The authors are particularly grateful to the anonymous users of Reddit whose data feature in this year's shared task dataset, to the clinical experts who annotated the data, to the American Association of Suicidology in making the dataset available, to the CLPsych 2024 shared task organizers. This work was supported by Sophia University Special Grant for Academic Research.

## Appendix A. List of suicide risk phrases.

1. attempt suicide, attempted suicide, attempting suicide, attempts of suicide, suicide attempt, suicide attempts
2. commit suicide, committed suicide, committing suicide
3. consider suicide, considered suicide, considering suicide
4. want to die, wanted to die, don't want to live
5. end my life
6. hang myself, hanging myself, myself hanging
7. kill me, kill myself, killed myself, killing me, killing myself
8. means of suicide, ways of dying
9. shoot me, shooting me, shoot myself, shooting myself
10. suicide plan, plan suicide
11. suicide thoughts, think about suicide, thinking about suicide, thinking of suicide, thought of suicide, thoughts of suicide, suicidal thoughts, suicide thoughts

## Appendix B. Prefix and phrases for generating summary

| Prefix | Phrases |
|---|---|
| This person feels | pain, anxious, sad, angry, agitated, trapped, hopeless, empty, guilt, shame, helpless, worthless, enraged, alone, isolated, failure |
| This person is dealing with issues with | friend, girlfriend, boyfriend, family, brother, sister, father, mother |
| This person has a problem of | eating, money, drug, alcohol |
| This person is struggling with | depression, trauma |
| This person is experiencing | bullying, abused, raped |